\documentclass{article}

\PassOptionsToPackage{numbers, sort&compress}{natbib}

\usepackage{graphicx}

\usepackage[preprint]{conf_2024}

\usepackage[utf8]{inputenc} 
\usepackage[T1]{fontenc}    
\usepackage{hyperref}       
\usepackage{url}            
\usepackage{booktabs}       
\usepackage{amsfonts}       
\usepackage{nicefrac}       
\usepackage{microtype}      
\usepackage{xcolor}         
\usepackage{longtable}      
\usepackage{wrapfig}

\usepackage{amsmath}
\usepackage{amssymb}
\usepackage{mathtools}
\usepackage{amsthm}

\usepackage{nccmath}
\usepackage{makecell}

\usepackage[skip=5pt]{caption}
\usepackage{titlesec}
\titleformat{\subsubsection}[hang]{\bfseries}{\thesubsubsection}{1em}{}

\usepackage[capitalize,noabbrev]{cleveref}

\title{Scalable Numerical Embeddings for Multivariate Time Series: Enhancing Healthcare Data Representation Learning}

\author{Chun-Kai Huang$^{1}$ \And Yi-Hsien Hsieh$^{1}$ \And Ta-Jung Chien$^{1}$  \And Li-Cheng Chien$^{2}$ \And Shao-Hua Sun$^{1,2,3}$ \And Tung-Hung Su$^{4,5}$ \And Jia-Horng Kao$^{4,6}$ \And Che Lin$^{1,3,7,8,}$ \thanks{Corresponding author: \texttt{chelin@ntu.edu.tw}} \AND \\
$^{1}$ Graduate Institute of Communication Engineering, National Taiwan University (NTU)\\
$^{2}$ Data Science Degree Program, NTU \\
$^{3}$ Department of Electrical Engineering, NTU \\
$^{4}$ Division of Gastroenterology and Hepatology, NTU Hospital \\
$^{5}$ Hepatitis Research Center, NTU Hospital \\
$^{6}$ Graduate Institute of Clinical Medicine, College of Medicine, NTU \\
$^{7}$ Center for Advanced Computing and Imaging in Biomedicine, NTU \\
$^{8}$ Smart Medicine and Health Informatics Program, NTU
} 

\begin{document}

\maketitle

\begin{abstract}
Multivariate time series (MTS) data, when sampled irregularly and asynchronously, often present extensive missing values. Conventional methodologies for MTS analysis tend to rely on temporal embeddings based on timestamps that necessitate subsequent imputations, yet these imputed values frequently deviate substantially from their actual counterparts, thereby compromising prediction accuracy. Furthermore, these methods typically fail to provide robust initial embeddings for values infrequently observed or even absent within the training set, posing significant challenges to model generalizability. In response to these challenges, we propose \textbf{SCA}lable \textbf{N}umerical \textbf{E}mbedding (\textbf{SCANE}), a novel framework that treats each feature value as an independent token, effectively bypassing the need for imputation. SCANE regularizes the traits of distinct feature embeddings and enhances representational learning through a scalable embedding mechanism. Coupling SCANE with the Transformer Encoder architecture, we develop the \textbf{S}calable n\textbf{UM}erical e\textbf{M}bedd\textbf{I}ng \textbf{T}ransformer (\textbf{SUMMIT}), which is engineered to deliver precise predictive outputs for MTS characterized by prevalent missing entries. Our experimental validation, conducted across three disparate electronic health record (EHR) datasets marked by elevated missing value frequencies, confirms the superior performance of SUMMIT over contemporary state-of-the-art approaches addressing similar challenges. These results substantiate the efficacy of SCANE and SUMMIT, underscoring their potential applicability across a broad spectrum of MTS data analytical tasks. 
  
\end{abstract}

\section{Introduction}
\label{sec:intro}
  Multivariate time series (MTS) data, a collection of observations recorded at discrete temporal intervals and comprising multiple interrelated variables, is pivotal across various sectors, including healthcare, energy, environmental science, and industrial monitoring.
  Unlike cross-sectional datasets, MTS data are inherently more informative, provided they are structured optimally—each variable is fully observed at consistent, isometric timestamps. 
  Recent advancements in this field have seen significant contributions from numerous studies \citep{sagheer_time_2019,che_recurrent_2016,zerveas_transformer-based_2021,liang_airformer_2022,wen_transformers_2023}, each focusing on harnessing the rich informational content inherent in MTS to improve prediction outcomes.

  However, most real-world MTS data are irregularly and asynchronously sampled. The irregularity causes the interval between two adjacent timestamps, and the total number of intervals varies. Not all feature variables are observed for each timestamp, creating data with a high missing rate. Traditionally, people first impute the missing values via statistic-based \citep{little_statistical_nodate} or learning-based models \citep{mattei_miwae_2019,du_saits_2023,kim_probabilistic_2023, zhao_transformed_2023} to obtain timestamp embeddings to these variables \citep{hochreiter1997long,breiman2001random,chung2014empirical,chen2016xgboost,vaswani_attention_2017}. In such scenarios, imputation is inevitable. Yet, the rationale behind imputation is complicated and challenging to justify in some domains, such as medicine and healthcare. Imputations that make sense to physicians may not work well for the learning model, while methods that learn well may not convince medical experts.
  
\begin{wrapfigure}{r}{0.5\textwidth}
  \begin{center}
    \includegraphics[width=0.48\textwidth]{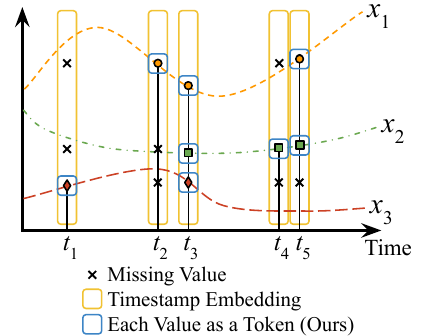}
  \end{center}
  \caption{\textbf{Embedding Multivariate Time Series Values.} The figure illustrates irregularly and asynchronously sampled MTS data with three variables ($x_1$-$x_3$) and five timestamps ($t_1$-$t_5$). The x marks represent missing values, and colored dots are observations. "Each Value as A Token (EVAT)" only embeds observations and bypasses missing values.}
  \label{fig:EVAT}
\end{wrapfigure}

The process of deriving robust representations from real-world multivariate time series (MTS) data introduces several intricate challenges. A critical issue is the need for distinct embeddings for tokens that, although numerically identical, represent different feature types. If not properly differentiated, these tokens can lead to confusion in downstream modules. Another issue is the inherent limitation in representing a continuous numerical value with a finite set of tokens.
Traditional approaches often employ quantization strategies to manage this issue \citep{9440752}. However, the effectiveness of such methods is generally inconsistent, largely contingent upon the chosen resolution of quantization, which can significantly impact performance.
Additionally, for embeddings associated with rarely observed or entirely unseen feature values, the quality representation learning often remains significantly suboptimal. This deficiency can adversely affect the overall effectiveness of the model. Ensuring high-quality embeddings under such conditions is crucial for maintaining the accuracy and generalizability of the predictive outcomes derived by the model, especially with a limited amount of training samples.

  To overcome the aforementioned challenges in learning representations from real-world MTS data, we introduce a novel methodology, \textbf{SCAlable Numerical Embedding} (\textbf{SCANE}). This approach innovatively leverages  
  both the feature type and the numerical value for representation learning, enhancing the model’s ability to handle diverse data attributes effectively.

SCANE capitalizes on the emerging “each value as a token (EVAT)” paradigm (see \cref{fig:EVAT}), which treats each numerical value as an independent token akin to word tokens in natural language processing tasks \citep{mikolov_efficient_2013}. This strategy facilitates the omission of imputations for missing values, streamlining the data preprocessing phase. Additionally, SCANE assigns a unique embedding to each feature type and adjusts this embedding by scaling it according to the observed values. 
This introduces a regularization mechanism that confines the infinite possibilities of embeddings, dictated by the continuous numerical values for each feature type, to a single embedding vector per feature type. The feature type determines the direction of the embedding, and the numerical value modulates its magnitude. Given that practical training data usually have a finite or even limited sample size, this strategic regularization simplifies the model’s embedding space and is expected to enhance interpretability and generalization ability.

Thus, by implementing these strategies, SCANE aims to provide more accurate initial embeddings, particularly for rarely observed or entirely unseen feature values, thus improving prediction accuracy and model robustness. The technical specifics and the operational framework of SCANE will be elaborated further in Section \ref{sec:scane}.



  To form a complete classifier, we take the Transformer Encoder \citep{vaswani_attention_2017} with SCANE to build \textbf{Scalable nUMerical eMbeddIng Transformer} (\textbf{SUMMIT}), followed by fully connected layers. With the inherent masking mechanism in the Transformer Encoder and the help of SCANE, we can perfectly mask all missing values. Through SCANE, SUMMIT is truly imputation-free and potentially benefits from better representations through the scalable embedding design. It not only distills the information from the observed value without interference from the imputation but also freely interacts with variables across temporal and feature-wise dimensions. Also, due to the attention mechanism and SCANE, we can visualize the attention map regarding all variables at every timestamp. This mitigates the difficulties of interpretability in our model, which is another important issue in domains such as healthcare. To simulate the whole attention flow in SUMMIT more closely, we revise the rollout attention \citep{abnar_quantifying_2020}. It helps us reveal the black box in our model and lets us know which feature the model emphasizes.

  We evaluate and compare SUMMIT with renowned and state-of-the-art (SOTA) models in this domain. Since medical data is a classic type of MTS with a high missing rate, experiments are conducted on three distinct electronic health record (EHR) datasets, one for chronic illness prediction and the other two for acute illness prediction. Experiment results provide preliminary evidence of SUMMIT's superiority. Moreover, the emphasized features indicated by the visualized revised rollout attention are consistent with the ones regarded as crucial in related medical literature.

\section{Related Work}
\label{sec:related_work}
  Sequence-to-sequence models such as gated recurrent unit (GRU) \citep{chung2014empirical}, Long-Short-Term Memory (LSTM) \citep{hochreiter1997long}, and Transformer-based models \citep{vaswani_attention_2017} have been widely used for MTS data \citep{ma_cdsa_2019,xu_anomaly_2022,zuo_svp-t_2023,grigsby_long-range_2023}. \citeauthor*{che_recurrent_2016} proposed GRU-D, a GRU-based model containing an imputation module, to handle the MTS data from the healthcare domain. \citeauthor*{sagheer_time_2019} proposed a Deep LSTM architecture model (DLSTM) to forecast petroleum production. More recent works have focused on Transformer-based models \citep{vaswani_attention_2017}. One prominent instantiation is the Time Series Transformer (TST), which proposes a Transformer-based framework for MTS representation learning \citep{zerveas_transformer-based_2021}. Additionally, \citeauthor{wu_deep_2020} have employed a Transformer encoder-decoder architecture for forecasting influenza prevalence, highlighting the superior performance of Transformer models compared to other deep learning and statistical models in forecasting tasks.

  Notably, deep learning models \citep{choi_learning_2020,zhang_graph-guided_2022,zhang_improving_2023} have found their popularity in the healthcare domain dealing with EHR data. Deep STI \citep{deep_STI_2021} proposed an RNN-VAE-based model to impute missing values in the EHR data. Deep STI can outperform traditional machine learning algorithms and statistical algorithms on the hepatocellular carcinoma (HCC) prediction task with the GRU-based classifier. These studies have promoted further development of deep learning in healthcare data analytics. Unlike the above works handling MTS data with imputations, recent works leveraging the EVAT-like concept, such as mTAN \citep{shukla_multi-time_2021} and STraTS \citep{tipirneni2022self}, achieved SOTA performance without the need for a separate imputation step on popular open MTS datasets with a high missing rate. 
  Our methodology extends the EVAT concept through a scalable embedding mechanism designed specifically to learn more effective representations for MTS. In our forthcoming analysis, we will benchmark SCANE/SUMMIT against these SOTA models to underscore their effectiveness and unique contributions in addressing similar challenges in the field.

\section{Methodology}
\label{sec:method}
This section employs the following notations: $\boldsymbol{X}$, is an $m \times n$ matrix, representing a time series data irregularly and asynchronously sampled and comprising $n$ features at $m$ timestamps. In this matrix, missing values are denoted by $\mathrm{Nan}$. The input time series data are summarized into a $k \times \left(n+1\right)$ matrix, $\boldsymbol{X}'=\begin{bmatrix}x'_{i,j}\end{bmatrix}$, following the procedure in Appendix \ref{appendix:summarization}, where $k$ is the number of evenly distributed summarization segments along the temporal axis and $n+1$ is for the $n$ features plus an additional feature counting the timestamps included in a summarization window. The first subscript of $x'_{i, j}$ is the temporal index of the summarization window, and the second subscript is the feature indicator. We defined a $k \times \left(n+1\right)$ missing mask matrix $\boldsymbol{\mathit{M}} = \begin{bmatrix}
    m_{i,j}
\end{bmatrix}$ to indicate the entry that is \textit{not} missing in $\boldsymbol{\mathit{X}}'$:
\begin{equation*}
m_{i,j} = 
\begin{cases}
     0\ ,\text{ if }  x'_{i, j} \text{ is missing.}\\
     1\ ,\text{ otherwise}.
\end{cases}
\end{equation*}

\subsection{Scalable Numerical Embedding}
\label{sec:scane}
\begin{figure}
    \centering\centerline{\includegraphics[width=.7\columnwidth]{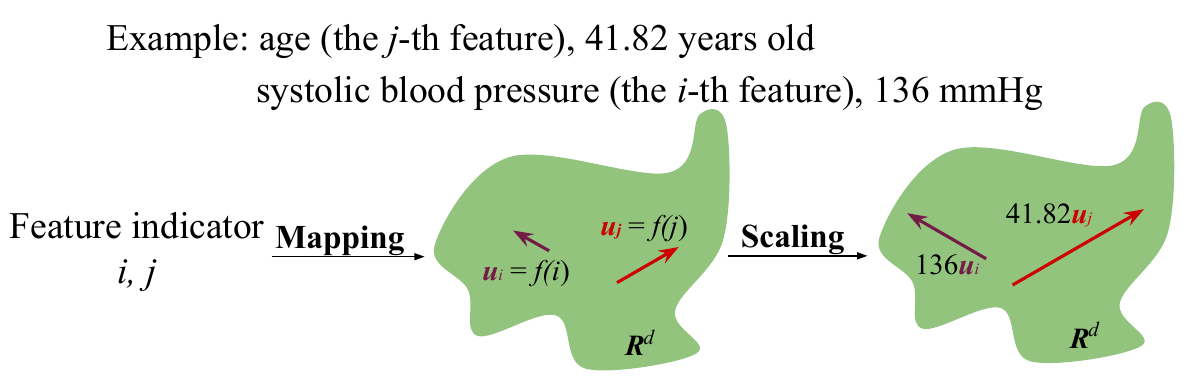}}
    \caption{\textbf{Scalable Numerical Embedding}: We take the value "age 41.82 years old" as an example to explain the process of SCANE. The process of SCANE can be divided into two steps: mapping and scaling. First, the "age" feature type indicator is mapped to the feature embedding $\boldsymbol{u}_j$ via a learnable function $f$. Second, we scale this feature embedding $\boldsymbol{u}_j$ according to its observed value. When the value is missing, the value will be assigned to a zero vector.}
    \label{fig:SCANE_concept}
\end{figure}
To extend the concept of EVAT and learn better representations, we propose SCANE that incorporates both the feature type and its observed value for embedding learning. It first maps the feature type indicator to a target vector space $\boldsymbol{\mathrm{U}}$ with dimension $d$. We call these assigned vectors "feature embedding." SCANE then scales feature embedding with each variable's observed value. The scaling design avoids the challenge of precision resolution issues as observed in \citep{jin2024floating}. \cref{eq:SNE_single} illustrates how SCANE embeds a single variable into a vector. 
\begin{equation}\label{eq:SNE_single}
\medmath{\mathrm{SCANE}\left(x'_{i,j},\ m_{i,j}\right) = \left(x'_{i,j}\cdot m_{i,j}\right) f\left(j\right) = \left(x'_{i,j} \cdot m_{i,j}\right) \boldsymbol{u}_j},
\end{equation}
where $f: \mathbb{N}\to{\boldsymbol{\mathrm{U}}}$ is realized through a single linear layer different for each feature, and $\boldsymbol{u}_j$ is feature $j$'s feature embedding $\in \boldsymbol{\mathrm{U}}$. SCANE assigns missing values to zero vector $\mathbf{0}^d$. We do not put any restrictions on the feature embeddings. The direction and the length of the feature embeddings are updated according to the training data. It is entirely data-driven. 
\cref{fig:SCANE_concept} gives an example to explain how SCANE works.

To generalize SCANE to its matrix form, we have:
\begin{multline*}
    \mathrm{SCANE}\left(\boldsymbol{X}',\ \boldsymbol{M}\right)=\\
    \begin{bmatrix}
    x'_{1,1} m_{1,1} \boldsymbol{u}_1 & x'_{1,2} m_{1,2} \boldsymbol{u}_2 & \dots & x'_{1,n+1} m_{1,n+1} \boldsymbol{u}_{n+1} \\
    x'_{2,1} m_{2,1} \boldsymbol{u}_1 & x'_{2,2} m_{2,2} \boldsymbol{u}_2 & \dots & x'_{2,n+1} m_{2,n+1} \boldsymbol{u}_{n+1} \\
    \vdots & \vdots & \ddots & \vdots \\
    x'_{k,1} m_{k,1} \boldsymbol{u}_1 & x'_{k,2} m_{k,2} \boldsymbol{u}_2 & \dots  & x'_{k,n+1} m_{k,n+1} \boldsymbol{u}_{n+1}
    \end{bmatrix}\ .
\end{multline*}
$\mathrm{SCANE}\left(\boldsymbol{X}',\ \boldsymbol{M}\right)$ is an $k \times \left(n+1\right) \times d$ tensor. Every feature embedding in the $ \mathrm{SCANE}\left(\boldsymbol{X}',\ \boldsymbol{M}\right) $ is scaled by the corresponding observed values.

Compared to the embedding method in common NLP applications, SCANE does not need to worry about the precision of the numerical value. For example, the embedding method in NLP may disassemble the embedding target "age 41.8243… years old" into "age + 4 + 1 + 8 + 2 + …". We do not know how many digits should be included to embed the value in this approach. Moreover, there may be unique values rarely seen in the dataset or even only shown in the external validation data. SCANE's scalable concept can regularize the underlying embedding mechanism via relatively stable embeddings learned (such as feature embeddings based on features with a clear characteristic or relationship) and provide better (initial) representations, not only for rare or unseen values but also for general observations with complex meanings. We regard this as SCANE's most valuable advantage.

\subsection{Transfomrer Encoder with Scalable Numerical Embedding}
\label{sec:transformer_SCANE}
We flatten and transpose $\mathrm{SCANE}\left(\boldsymbol{X}',\ \boldsymbol{M}\right)$ into a $k\left(n+1\right) \times d$ vector. Similarly, we flatten the matrix $\boldsymbol{M}$ into a $1 \times k\left(n+1\right)$ vector. That is,

\begin{align*}
\Bar{\boldsymbol{X}} &= \left(\mathrm{flatten}\left(\mathrm{SCANE}\left(\boldsymbol{X}',\boldsymbol{M}\right)\right)\right)^T \\
    &= \medmath{\begin{bmatrix}
        x_{1,1} m_{1,1} \boldsymbol{u}_{1} & x_{1,2} m_{1,2} \boldsymbol{u}_{2} & \dots & x_{k,n+1} m_{k,n+1} \boldsymbol{u}_{n+1}
    \end{bmatrix}^T}\\ 
    \Bar{\boldsymbol{M}} &= \mathrm{flatten}\left(\boldsymbol{M}\right) =\begin{bmatrix}
    m_{1,1} & m_{1,2} & \dots & m_{k,n+1}
    \end{bmatrix}\ .
\end{align*}
In the Transformer Encoder's self-attention module, we take $\boldsymbol{Z}=\Bar{\boldsymbol{X}}+PE$ to obtain the query $\boldsymbol{Q} = \boldsymbol{Z}\boldsymbol{W_q}$, the key $\boldsymbol{K} = \boldsymbol{Z}\boldsymbol{W_k}$, and the value $\boldsymbol{V} = \boldsymbol{Z}\boldsymbol{W_v}$. $PE$ is the positional encoding. The $\boldsymbol{W_q}, \boldsymbol{W_k}, \boldsymbol{W_v} \in \mathbb{R}^{d \times d}$ are learnable weights. To avoid paying attention to missing values, we use the masking mechanism \citep{vaswani_attention_2017} to mask them in $\boldsymbol{Z}$. This can only be realized under the EVAT concept.

\begin{multline}
\label{eq:masking_attention}
    \mathrm{Attention}\left(\boldsymbol{Q},\boldsymbol{K},\boldsymbol{V},\Bar{\boldsymbol{M}}\right) =\\
    \resizebox{0.7\columnwidth}{!}{$\medmath{\mathrm{softmax}\left(\lim_{a\to-\infty}\left( \frac{\left(\boldsymbol{a} \right)^{k\left(n+1\right) \times 1} \left(\mathbf{1}^{1 \times k\left(n+1\right)}-\Bar{\boldsymbol{M}}\right) + \left(\boldsymbol{Q}\boldsymbol{K}^T\right)}{\sqrt{d}}\right)\right)\boldsymbol{V},}
    $}
\end{multline}
where $d$ is the dimension of embeddings as a suggested scaling factor \citep{vaswani_attention_2017} and $a$ is a number approach negative infinity. $\mathbf{1}^{1 \times k\left(n+1\right)}$ is an $1 \times k\left(n+1\right)$ matrix whose entries are all 1. The vector $\boldsymbol\lim_{a\to-\infty}{(\boldsymbol{a})}^{k\left(n+1\right) \times 1}$ is $k\left(n+1\right) \times 1$ and its entries all equal $a$. All attention weights with missing values as a key will be suppressed by the number $a$, which approaches negative infinity and will be zero after $\mathrm{softmax}$. \cref{eq:masking_attention} shows how to use the mask to avoid paying attention to missing values in the self-attention module with SCANE.

We can mask missing values independently due to EVAT. If we apply timestamp embedding instead, the missing values would be bound together with other non-missing values. It is impossible to avoid imputation in this case. The missing value would be assigned to a zero vector in SCANE, and its contextual embedding from the second Transformer Encoder stack would comprise other non-missing values’ embeddings. So, we only mask missing values in the first layer in the entire Transformer Encoder stacks.

\subsection{Revised Rollout Attention}
\label{sec:revised_rollout}
Rollout attention is a method to simulate the information flow across all attention modules \citep{abnar_quantifying_2020}. While Transformer-based models often have multiple encoder stacks, rollout attention performs matrix multiplication on all attention matrices from the stacks to thoroughly evaluate how an input is attended. We proposed a revised version of rollout attention to better address input with significant missing values, detailed in Appendix \ref{appendix:rolloutatt}


\section{Experiment and Result}
\label{sec:ext_and_result}

\subsection{Datasets}
Inspired by our major benchmarks \citep{shukla_multi-time_2021, tipirneni2022self}, we conduct experiments on three distinct EHR datasets: MIMIC-III (public), PhysioNet2012 (public), and the Anonymous Hospital Hepatocellular Carcinoma Dataset (private). All three datasets originate from the healthcare domain and are characterized by irregular sampling and unsynchronized measurement, thereby presenting challenges for MTS binary classification tasks.

\subsubsection{MIMIC-III (MI3)}
This public dataset comprises numerous ICU patients with laboratory test results, encompassing 128 numerical features (e.g., heart rate, oxygen saturation, fraction inspired oxygen) and 3 categorical features (e.g., Glasgow coma scale eye-opening), as detailed in Appendix \ref{appendix:features}. Preprocessing and train-test-split of the data follows \citeauthor{tipirneni2022self}'s work. The binary classification task for this dataset entails predicting patients' survival during their hospital stay. The observation window for this task spans 48 hours after patients' initial hospitalization, with the summarization window length ($p$) set to 2 hours. The dataset comprises 44812 samples, consisting of 5150 positive and 39662 negative samples, with an imbalance ratio of 0.130. After summarization, the average missing rate of all features in the summarized data is 0.8814.

\subsubsection{PhysioNet2012 (P12)}
The public dataset is derived from the 2012 PhysioNet challenge \citep{goldberger_physiobank_2000}, encompassing 11988 intensive care unit (ICU) stays lasting at least 48 hours. The central task for this dataset is to predict if the patient dies during their hospital stay. The dataset exhibits class imbalance with 1707 positive samples and 10281 negative samples, with an imbalance ratio of 0.142. The dataset consists of 40 numerical features (e.g., glucose, urine, cholesterol) and 2 categorical features (e.g., sex and ICU type), detailed in Appendix \ref{appendix:features}. The observation window spans 48 hours, with a summarization window length ($p$) set to 2 hours. After summarization, there are 24 summarization windows, and the average missing rate of all features is 0.7377. We adopt the train-test-split strategy from \citeauthor{horn_set_2020}.

\subsubsection{Hepatocellular Carcinoma Dataset (HCC)}
This private dataset is sourced from the Anonymous Hospital and comprises records from patients over a one-year-length observation window since patients' first diagnosis record. The dataset includes 30 numerical features (e.g., alanine aminotransferase, alpha-fetoprotein, prothrombin time) and 8 categorical features (e.g., Anti-HCV, Anti-HBe, HBsAg), which are listed in Appendix \ref{appendix:features}. The primary objective of this dataset is to predict whether a patient will develop hepatocellular carcinoma within the ensuing five years. The dataset exhibits a pronounced class imbalance with 1523 positive and 32773 negative samples, indicating an imbalance ratio of 0.046. After the summarization with a summarization window length ($p$) of 90 days, the average missing rate of all features amounts to 0.7464, a remarkably high proportion of missing values. We perform a stratified train-test-split for model evaluation to divide samples into training and testing sets with a ratio of 8:2.

\subsection{Models}
Several baseline models were selected to compare against the proposed SUMMIT. The non-sequential benchmarks encompass Random Forest \citep{breiman2001random}, and XGBoost \citep{chen2016xgboost}, while the deep-learning-based benchmarks include the basic Transformer Encoder \citep{vaswani_attention_2017}, Temporal Convolutional Network (TCN) \citep{bai2018empirical}, Simply Attend and Diagnose (SAnD) \citep{song2018attend}, MultiTime Attention Networks (mTAN) \citep{shukla_multi-time_2021}, and Self-supervised Transformer for Time-Series (STraTS) \citep{tipirneni2022self}. These models have garnered widespread use in many areas, such as the healthcare sector. The detailed settings of each model, including the architecture and hyperparameters used, are shown in Appendix \ref{appendix:grid_search}. We take each variable's global mean and mode from the training set to impute the training and testing sets for models without any genuine design to deal with missing values.

Our proposed model, SUMMIT, employs a Transformer Encoder with SCANE as the feature extractor, followed by fully connected layers to form the classifier. Since we have both numerical and categorical features, we form two separate SCANE modules for each.

\subsection{Experimental Settings}
\label{sec:setup}
We list the hardware and platform used in Appendix \ref{appendix:plf_info}). For baseline models, we adopt the best settings reported in the original papers or package documentation. Detailed settings are shown in Appendix \ref{appendix:grid_search}. For SUMMIT, we applied the following settings. Given the inherent class imbalance in the datasets, we adopt focal loss \citep{lin_focal_2018}. To optimize SUMMIT's performance, we employ the following grid search strategy. We select hyperparameters according to models' performance on the validation set, which constitutes 20\% of the training set. Throughout the training process, we monitor the model's performance on the validation set every 5 epochs and halt the process if there is no improvement in the area under the precision-recall curve (AUPRC) or the area under the receiver operating characteristic curve (AUROC) for a continuous span of 30 epochs. The batch size is fixed at 256 for all experiments. The maximal training epochs for the MI3, P12, and HCC are set to 400, 500, and 100, respectively, ensuring adequate training for each dataset to capture underlying patterns and achieve convergence. The resulting hyperparameters are detailed in \ref{appendix:grid_search}.

\subsection{Metrics}
Because all of the datasets we used are imbalanced, we take AUPRC as the primary metric to evaluate each model's performance. AUPRC is more indicative of an imbalanced binary classification task than AUROC \citep{saito_precision-recall_2015}. We also adopt AUROC and concordance index (c-index) as auxiliary metrics. However, the event time is not contained in P12 and MI3, so the c-index is substituted for accuracy on these two datasets to evaluate models. The decision threshold to calculate accuracy here is set at $0.5$.

\subsection{Results and Discussion}
\subsubsection{Overall Result}
\label{sec:overall}
\cref{tab:overall} depicts the performance of baseline models and our model, SUMMIT, on the MI3, P12, and HCC, respectively. Our model, SUMMIT, outperforms other models in all datasets on AUPRC, which is the primary metric. It also performs well in other auxiliary metrics, achieving the best and second-best values in terms of AUROC and accuracy on the P12. With this result, we can confirm that it can learn well from data featuring high missing rates without the need for imputation. SUMMIT faithfully learns from what we observed and achieved SOTA performance on the main metric, AUPRC. This supports the idea that SCANE is a promising choice for irregular and asynchronous MTS data.

\begin{table*}[h]
    \centering
    \resizebox{\linewidth}{!}{
    \begin{tabular}{cccccccccc}
    \toprule
    \textbf{Dataset}             & \multicolumn{3}{c}{\textbf{MI3}} & \multicolumn{3}{c}{\textbf{P12}} & \multicolumn{3}{c}{\textbf{HCC}}  \\ \cmidrule{1-1} \cmidrule(lr){2-4} \cmidrule(lr){5-7} \cmidrule(lr){8-10}
    \textbf{Metric}              & \textbf{AUPRC}  & \textbf{AUROC} & \textbf{accuracy} & \textbf{AUPRC} & \textbf{AUROC} & \textbf{accuracy} & \textbf{AUPRC} & \textbf{AUROC} & \textbf{c-index} \\ \midrule
    \textbf{Random Forest}       & 
    \makecell{0.4367\\$\pm$0.0517}       &    \makecell{0.8319\\$\pm$0.0209}       &    \makecell{0.8965\\$\pm$0.0105} &
    \makecell{0.4805\\$\pm$0.0533}       &    \makecell{0.8270\\$\pm$0.0228}       &    \makecell{0.8663\\$\pm$0.0146} & 
    \makecell{0.3934\\$\pm$0.0583}       &    \makecell{0.8705\\$\pm$0.0232}       &    \makecell{0.8637\\$\pm$0.0227}  
    \\ \midrule
    \textbf{XGBoost}             & 
    \makecell{0.4553\\$\pm$0.0527}       &    \makecell{0.8247\\$\pm$0.0209}       &    \makecell{0.8968\\$\pm$0.0105} &
    \makecell{0.4980\\$\pm$0.0544}       &    \makecell{0.8453\\$\pm$0.0203}       &    \makecell{0.8708\\$\pm$0.0140} & 
    \makecell{0.3887\\$\pm$0.0592}       &    \makecell{0.8714\\$\pm$0.0215}       &    \makecell{0.8644\\$\pm$0.0209}  
    \\ \midrule
    \textbf{Transformer Encoder} &   
        \makecell{0.5074\\$\pm$0.0510}       &    
    \makecell{0.8606\\$\pm$0.0187}       &    \makecell{0.8953\\$\pm$0.0105} &  
    \makecell{\underline{0.5435}\\$\pm$0.0560}       &
    \makecell{0.8572\\$\pm$0.0200}       &   
    \makecell{0.8767\\$\pm$0.0131} & 
    \makecell{0.4139\\$\pm$0.0571}       & 
    \makecell{\textbf{0.8964}\\$\pm$0.0171}       &  
    \makecell{\textbf{0.8888}\\$\pm$0.0171} 
    \\ \midrule
    \textbf{TCN}                 & 
    \makecell{0.5128\\$\pm$0.0377}       &    \makecell{0.8734\\$\pm$0.0165}       &    \makecell{0.8999\\$\pm$0.0098} &
    \makecell{0.4725\\$\pm$0.0494}       &    \makecell{0.8272\\$\pm$0.0263}       &    \makecell{0.8581\\$\pm$0.0134} &  
    \makecell{0.3725\\$\pm$0.0661}       &    \makecell{0.8684\\$\pm$0.0493}       &    \makecell{0.8616\\$\pm$0.0187} 
    \\ \midrule
    \textbf{SAnD}                 & 
    \makecell{0.5463\\$\pm$0.0462}       &    \makecell{0.8774\\$\pm$0.0096}       &    \makecell{0.9023\\$\pm$0.0123} &
    \makecell{0.4615\\$\pm$0.0598}       &    \makecell{0.8227\\$\pm$0.0245}       &    \makecell{0.8674\\$\pm$0.0179} &  
    \makecell{0.3769\\$\pm$0.0337}       &    \makecell{0.8836\\$\pm$0.0090}       &    \makecell{0.8763\\$\pm$0.0087} 
    \\ \midrule
    \textbf{mTAN}                 & 
    \makecell{0.5536\\$\pm$0.0359}       &    \makecell{0.8826\\$\pm$0.0163}       &    \makecell{0.9037\\$\pm$0.0227} &
    \makecell{0.4991\\$\pm$0.0521}       &    \makecell{0.8444\\$\pm$0.0267}       &    \makecell{\textbf{0.8863}\\$\pm$0.0127} & 
    \makecell{\underline{0.4545}\\$\pm$0.0264}       &    \makecell{0.8762\\$\pm$0.0135}       &    \makecell{0.8466\\$\pm$0.0138} 
    \\ \midrule
    \textbf{STraTS}                 & 
    \makecell{\underline{0.5886}\\$\pm$0.0546}       &    \makecell{\underline{0.8936}\\$\pm$0.0021}       &    \makecell{\underline{0.9044}\\$\pm$0.0104} &
    \makecell{0.5206\\$\pm$0.0534}       &    \makecell{\underline{0.8596}\\$\pm$0.0224}       &    \makecell{0.8253\\$\pm$0.0135} &  
    \makecell{0.4270\\$\pm$0.0186}       &    \makecell{\underline{0.8963}\\$\pm$0.0088}       &    \makecell{\textbf{0.8888}\\$\pm$0.0086} 
    \\ \midrule
    \textbf{SUMMIT}           & 
    \makecell{\textbf{0.6328}\\$\pm$0.0277}       &    \makecell{\textbf{0.9035}\\$\pm$0.0092}       &    \makecell{ \textbf{0.9111}\\$\pm$0.0060} &
    \makecell{\textbf{0.5504}\\$\pm$0.0563}       &    \makecell{\textbf{0.8602}\\$\pm$0.0197}       &    \makecell{\underline{0.8783}\\$\pm$0.0129} & 
    \makecell{\textbf{0.4553}\\$\pm$0.0577}       &    \makecell{0.8943\\$\pm$0.0179}       &    \makecell{\underline{0.8867}\\$\pm$0.0179} 
    \\ \bottomrule
    \end{tabular}%
    }
    \caption{This table shows the overall results of each model on the three test sets. We mark the best value in \textbf{boldface} and \underline{underline} the second-best value for each metric. The value in parentheses is the 95\% of the confidence interval of the 1000 bootstrap times in the test set.}
    \label{tab:overall}
\end{table*}

\begin{figure*}[htp]
    \centering
    \centerline{\includegraphics[width=\textwidth]{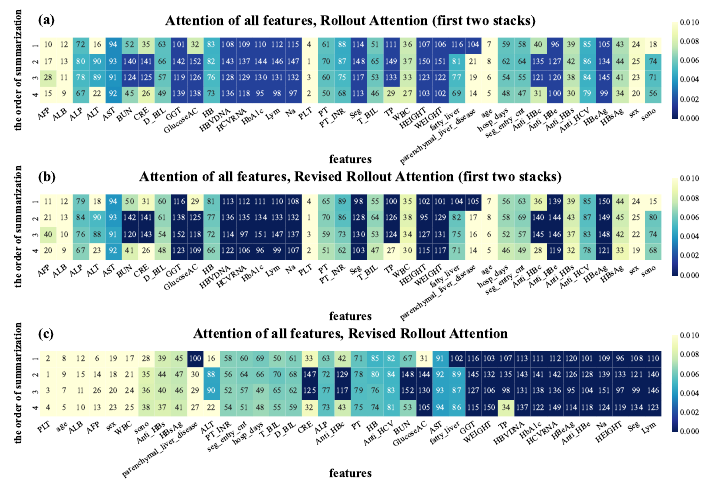}}
    \caption{\textbf{Attention Weights Visualization.} The number in each cell ranks the corresponding attention weights: the smaller, the higher. (a) Rollout Attention on the Attention Weights from the First Two Stacks of SCANE. (b) Revised Rollout Attention on the Attention Weights from the First Two Stacks of SCANE. (c) Revised Rollout Attention on all SUMMIT's Attention Stacks. We rearrange the columns based on each feature’s mean rank across the timestamps.}
    \label{fig:attn_map}
    \vspace{-0.4cm}
\end{figure*}

\subsubsection{Attention Weight Visualization}
We focus on a positive sample from the HCC for attention weight visualization, which is suggested by our partner medical expert as an example of further analysis. Given that we aggregate the output of the feature extractor through means, we compute the average of the $152\times152$ revised rollout attention (with $38$ variables $\times 4$ timestamps $=152$ features per sample under SCANE scenario) with respect to columns. We then reshape the resulting $1\times152$ matrix into a $4\times38$ matrix, where columns and rows represent features and summarization index, respectively. Every cell in this $4\times38$ matrix can be considered the feature importance. It is somehow the linear combination weight of the corresponding value $\boldsymbol{v}$.

We consider the first two stacks' attention weights to compare the difference between rollout attention and the revised rollout attention. \cref{fig:attn_map} (a) shows the rollout attention map. Without neglecting meaningless residual connections in the original version, the rollout attention tends to highlight the missing value and results in a smaller brightness disparity in the visualization. Moreover, the redundant residual connection in the rollout attention would blend the information flow from the missing and non-missing values. This causes the ranking to be inconsistent in the rollout attention and the revised rollout attention. The phenomenon can be observed in \cref{fig:attn_map} (a) and \cref{fig:attn_map} (b). For example, the rankings of each summarized ”AFP” (Alpha-Fetoprotein) are the 10th, the 17th, the 28th, and the 15th in the rollout attention. However, the rankings in the revised rollout attention are the 11th, the 21st, the 40th, and the 20th.

We further present this sample's revised rollout attention in \cref{fig:attn_map} (c). Notably, the top-5 features that significantly influence our model’s prediction for this sample are ”PLT” (platelet), ”age,” ”ALB” (albumin), ”AFP” (Alpha-Fetoprotein), and ”sex.” They are highly related to hepatocellular carcinoma \citep{pang_prognostic_2015,carr_serum_2017,bhangui_salvage_2016,sauzay2016alpha}. This result strongly supports SUMMIT’s interpretability, suggesting the potential for further investigation of its identified crucial factors from a medical perspective.  

\vspace{-0.3cm}

\subsubsection{Feature Embedding Visualization}

\begin{figure*}[htp]
    \centering
    \centerline{\includegraphics[width=0.9\textwidth]{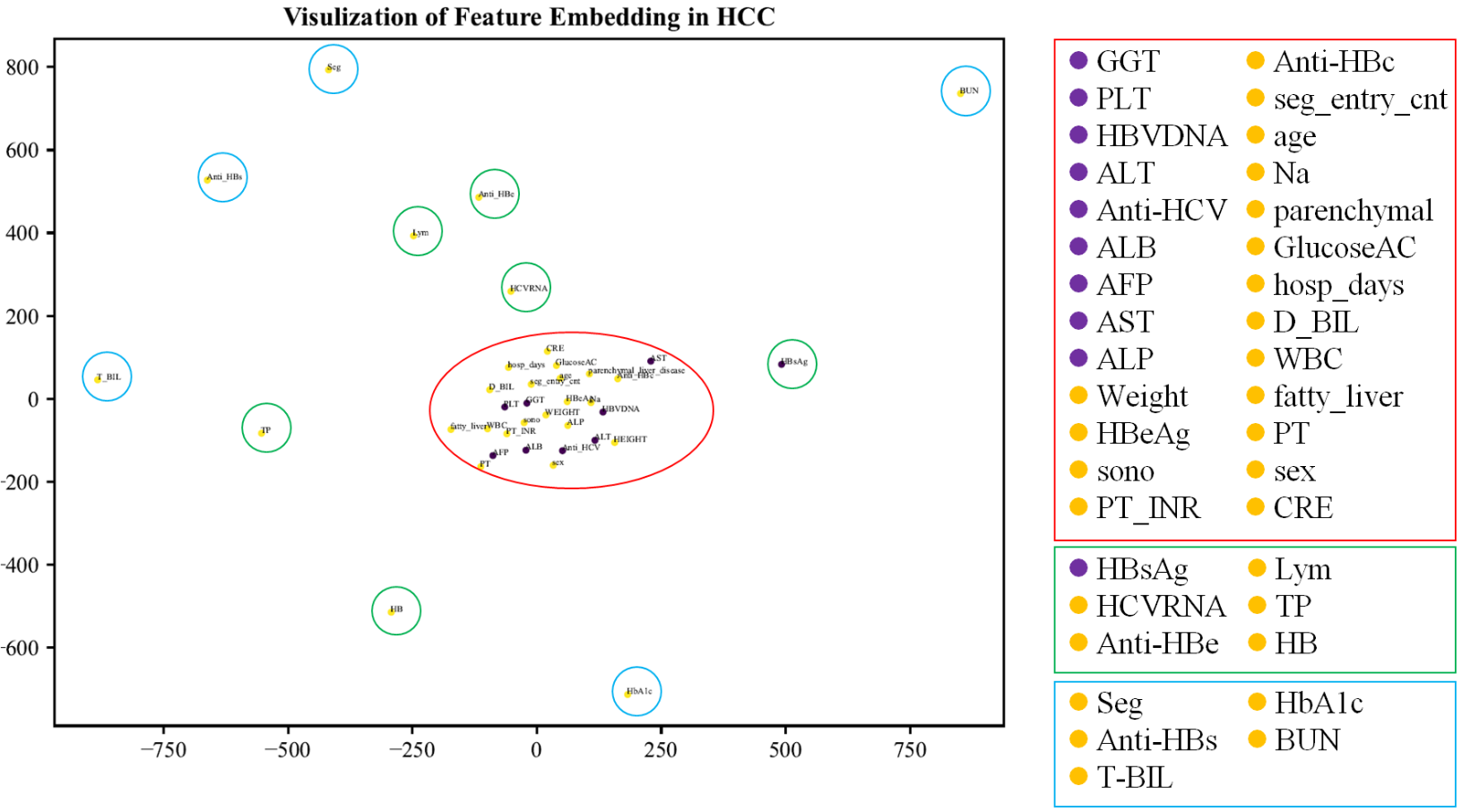}}
    \caption{\textbf{Feature Embedding Visulization}: We visualize the feature embedding from HCC in this plot. The purple dots represent the hepatocellular-carcinoma-related features suggested by our partner medical experts. For the full feature names, please refer to Appendix \cref{table:ntuh_feature}.}
    \label{fig:viz_feature_embedding}
    \vspace{-0.3cm}
\end{figure*}

We visualize the feature embeddings learned from the HCC dataset via t-SNE \citep{JMLR:v9:vandermaaten08a}. As illustrated in \cref{{fig:viz_feature_embedding}}, most feature embeddings (without scaling) belonging to features highly related to HCC development (verified by our partnering clinicians and are shown in purple dots) indeed appear close to each other. In this case, SCANE's scalable mechanism allows all these features' embeddings (even for those possible values beyond current observations) to maintain their relationship (e.g., angle, similarity), thus obtaining better initial representations under a limited amount of training data. In addition, we can observe other feature embeddings in the cluster (the yellow dots circled in red) formed by HCC-related feature embeddings. Such information can help identify novel relationships between features not yet well-studied in the medical domain.

\vspace{-0.2cm}
\subsubsection{Ablation Study}
\begin{figure}[htp]
    \centering\centerline{\includegraphics[width=0.8\columnwidth]{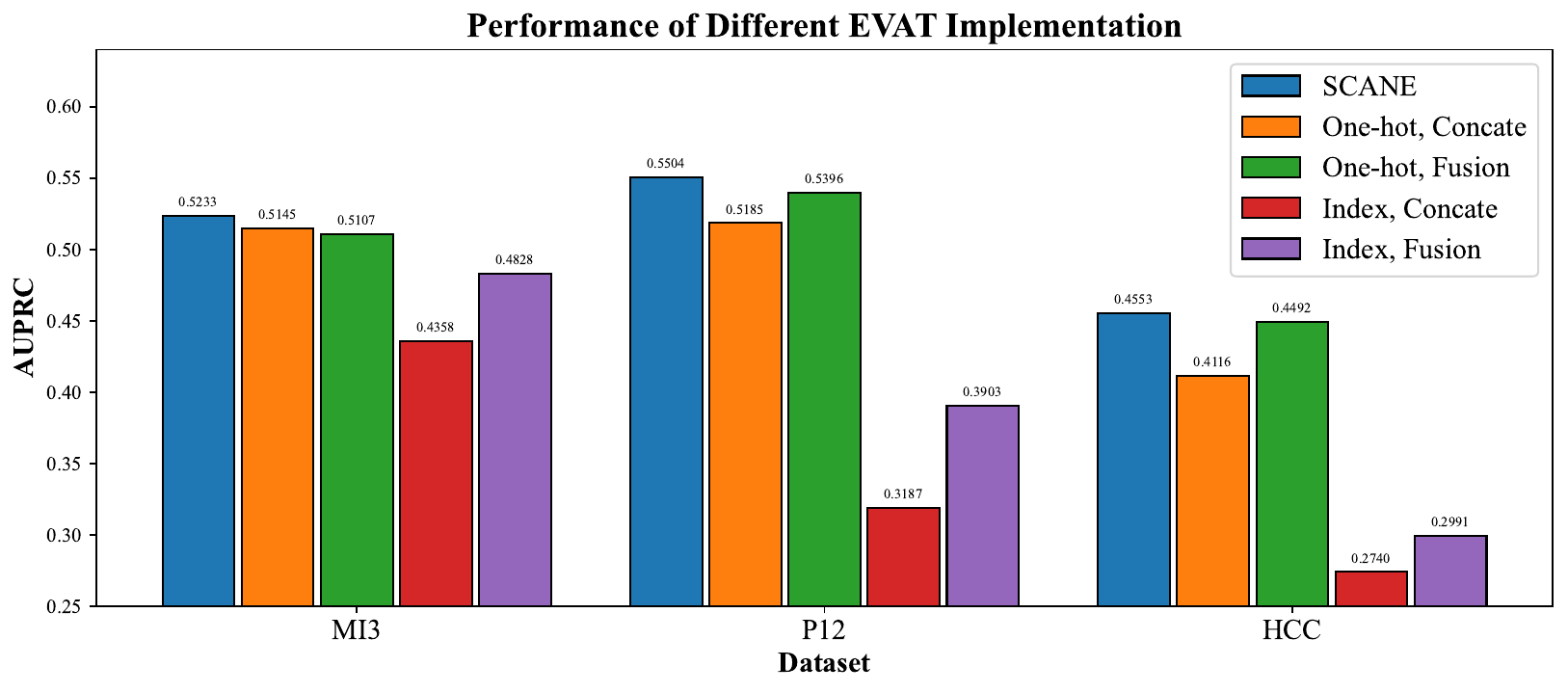}}
    \caption{\textbf{EVAT Implementation Comparison}: The setting in this ablation study follows \cref{sec:setup}. Intuitively, other than metric performance, the SCANE is also the most parameter-efficient implementation than other naive EVAT implementations.}
    \label{fig:EVAT_comparison}
\end{figure}
To further evaluate SCANE's efficacy, we implement other naive EVAT methods to replace the SCANE in SUMMIT. These naive implementations include "index, concate", "index, fusion", "one-hot, concate", and "one-hot fusion". "Concate" means that we concatenate the representation of the feature type indicator and its observed value together, then feed them into a linear layer to get $d$-dimensioned embeddings. "One-hot" differs from the "index" in terms of representing the feature type indicator in one-hot vectors. The terms postfixed with "fusion" pass the feature type indicator and the observed value to a dense layer and obtain the embedding with the same dimension as the dense layer's output. To avoid the embeddings of missing values carrying redundant information in the Transformer Encoder, we force these embeddings to $\mathbf{0}^d$. \cref{fig:EVAT_comparison} shows the performance of different EVAT implementations on the three datasets in our primary metric (AUPRC), and SCANE outperforms all other basic EVAT implementations.

\vspace{-0.2cm}
\subsubsection{Limitations}
Inspired by the works of the main baseline models compared, all datasets on which we conducted the experiments consisted of medical MTS. Although medical datasets are indeed classic examples of MTS with a high missing rate, and the characteristics of the three datasets are not completely duplicated (for instance, MI3 and P12 are for acute ICU data while HCC is for chronic disease records), evidence provided by the current results is still limited to the medical domain. Extended experiments shall be performed in future works to further support the work's goal.

\section{Conclusion}
\label{sec:conclusion}
In this work, we propose SCANE, a novel method that integrates the feature type and its observed value for representation learning. SCANE first leverages the EVAT concept to bypass potentially harmful missing value imputations. It then regularizes the embedding learning through a scaling mechanism and provides better initial embeddings, especially for rarely observed or even unseen feature values. We further combine SCANE with the Transformer Encoder to build SUMMIT, a fully functional, imputation-free classifier for practical applications. 
Our evaluation across three EHR datasets demonstrates SUMMIT's superiority over existing SOTA benchmarks. Furthermore, a revised rollout attention mechanism in SUMMIT enhances model interpretability, offering insights into its decision-making processes. Starting from the MTS of the medical domain, we expect SCANE/SUMMIT to serve as a powerful tool to address missing values in general MTS data and benefit from learning better representations through its scalable embedding design.

\bibliography{references}

\clearpage

\appendix
\onecolumn

\section{Summarization}
\label{appendix:summarization}
This section employs the following notations: $\boldsymbol{X}$, is an $m \times n$ matrix, representing a time series data irregularly and asynchronously sampled and comprising $n$ variables at $m$ timestamps. In this matrix, missing values are denoted by $\mathrm{Nan}$. We denote $t_i$ as the timestamp associated with the $i$-th row. The sequence of timestamps $\{t_i\}_1^m$ is arranged in ascending order. All input time series data are gathered within an observation window of length $T$. We denote $x_{i,j}$ as the entry of the $j$-th feature at the timestamp $t_i$.

To better control the number of timestamps of each sample and preliminarily alleviate the missing value issue, we follow \citeauthor{deep_STI_2021}'s work to apply the \emph{summarization} strategy on the input data. Specifically, given a summarization time duration $p$, we obtain $k\ (=\lfloor T/p\rfloor)$ summarization intervals. The formed summarization intervals are $\left[t_1,\ t_1+p\right)$, $\left[t_1+p,\ t_1+2p\right)$, ..., and $\left[t_1 + \left(k-1\right)p,\ t_1+T\right)$. We then assign the rows of $\boldsymbol{X}$ to the summarization intervals where their timestamps belong. Every summarization window uses the mean, the mode (the most frequently observed value), or the last observed value of the collected rows to represent the value of features in the interval. The mean is used to represent the numerical feature; the mode and the last observed value are used to represent the categorical feature. The mean and the mode are computed by dropping missing values. If there is no observation of a feature in the interval, it will assign $\mathrm{Nan}$ for this feature. This strategy also counts the number of rows in each summarization window and records it as an additional feature, "segment entry count," to the $\boldsymbol{X}$.

The input time series data are then summarized into a $k \times \left(n+1\right)$ matrix, $\boldsymbol{X}'=\begin{bmatrix}x'_{i,j}\end{bmatrix}$. The first subscript of $x'_{i, j}$ is the index of the summarization window, and the second subscript is the feature indicator. We defined a $k \times \left(n+1\right)$ missing mask matrix $\boldsymbol{\mathit{M}} = \begin{bmatrix}
    m_{i,j}
\end{bmatrix}$ to indicate the entry that is \textit{not} missing in $\boldsymbol{\mathit{X}}'$:
\begin{equation*}
m_{i,j} = 
\begin{cases}
     0\ ,\text{ if }  x'_{i, j} \text{ is missing.}\\
     1\ ,\text{ otherwise}.
\end{cases}
\end{equation*}

\section{Revised Rollout Attention}
\label{appendix:rolloutatt}
Suppose $\boldsymbol{W}_i$ is the attention weight from SUMMIT's $i$-th stacked attention module in \cref{eq:masking_attention}, and there are $N$ stacks in a Transformer Encoder. It defines the raw attention $\boldsymbol{A}_i=0.5\boldsymbol{W}_i+0.5\boldsymbol{I}$ to reflect the residual connection, where $\boldsymbol{I}$ is the identity matrix. The parameter $0.5$ is used to normalize the raw attention. With the raw attention from each attention module, the rollout attention $\Tilde{\boldsymbol{A}}$ is:
\begin{equation*}
    \Tilde{\boldsymbol{A}} = \boldsymbol{A}_N \cdot \boldsymbol{A}_{N-1} \cdot ... \cdot \boldsymbol{A}_2 \cdot \boldsymbol{A}_1\ .
\end{equation*}

The original rollout attention does not discard the missing values. In SUMMIT, it is affected by the residual connection of missing values' zero vectors that shall have no practical impact. We revise the raw attention $\boldsymbol{A}_1$ in $\Tilde{\boldsymbol{A}}$ to fit our application with significant missing values. The revised raw attention $\boldsymbol{A}_1'$ is defined as:
\begin{equation}\label{eq:rolloutA1_prime}
    \boldsymbol{A}_1' = \mathrm{norm}\left(\boldsymbol{W}_1 + \mathrm{diag}\left(\Bar{\boldsymbol{M}}\right)\right),
\end{equation}
where $\mathrm{norm}$ is the normalization function in terms of row and $\mathrm{diag}$ is an operator for constructing a diagonal matrix whose diagonal entries are $m_{1, 1}$, $m_{1, 2}$, $\dots$, $m_{k, \left(n\right)}$, and $m_{k, n+1}$. The modified rollout attention for our work becomes:
\begin{equation*}
    \Tilde{\boldsymbol{A}}' = \boldsymbol{A}_N \cdot \boldsymbol{A}_{N-1} \cdot ... \cdot \boldsymbol{A}_2 \cdot \boldsymbol{A}_1'\ .
\end{equation*}
Given that we only mask what is missing at the first stack of SCANE, we only need to revise the raw attention for the first attention module to block the weights corresponding to missing values. With this modification, the attention to missing value embeddings (zero vectors) will not be considered throughout the rollout attention propagation.

\section{Features in the Datasets}
\label{appendix:features}
Tables \ref{table:ntuh_feature} to \ref{table:mi3_feature} list the full feature set of the datasets applied. In \cref{table:ntuh_feature}, "fatty\_liver" is a categorical feature to show the fatty liver severity; "parenchymal\_liver\_disease" is also a categorical feature to represent the severity of cirrhosis; "hosp\_days" is the number of hospitalization days; "sono" represents whether a patient has the abdominal ultrasound imaging. In \cref{table:p12_feature}, "MechVent" means whether a patient uses mechanical ventilation in the ICU.

    \begin{longtable}{ccc}
    \toprule
    \centering
    \textbf{Feature}                                &   \textbf{Feature Type}       \\  \midrule
        Age                                         &   Numerical                   \\
        Gender                                      &   Numerical                   \\
        RR (Respiratory Rate)                       &   Numerical                   \\
        Weight                                      &   Numerical                   \\
        HR (Heart Rate)                             &   Numerical                   \\
        MBP (Mean Blood Pressure)                   &   Numerical                   \\
        DBP (Diastolic Blood Pressure)              &   Numerical                   \\
        SBP (Systolic Blood Pressure)               &   Numerical                   \\
        O$_{2}$ Saturation                          &   Numerical                   \\
        CRR (Capillary Refill Rate)                 &   Numerical                   \\
        Base Excess                                 &   Numerical                   \\
        Calcium Free                                &   Numerical                   \\
        Lactate                                     &   Numerical                   \\
        PCO$_{2}$                                   &   Numerical                   \\
        PO$_{2}$                                    &   Numerical                   \\
        Potassium                                   &   Numerical                   \\
        Total CO$_{2}$                              &   Numerical                   \\
        pH Blood                                    &   Numerical                   \\
        Glucose (Blood)                             &   Numerical                   \\
        Urine                                       &   Numerical                   \\
        Solution                                    &   Numerical                   \\
        Normal Saline                               &   Numerical                   \\
        FiO$_{2}$                                   &   Numerical                   \\
        ALP (Alkaline Phosphatase)                  &   Numerical                   \\
        ALT (Alanine Aminotransferase)              &   Numerical                   \\
        AST (Aspartate Aminotransferase)            &   Numerical                   \\
        Anion Gap                                   &   Numerical                   \\
        BUN (Blood Urea Nitrogen)                   &   Numerical                   \\
        Bicarbonate                                 &   Numerical                   \\
        Bilirubin (Total)                           &   Numerical                   \\
        Calcium Total                               &   Numerical                   \\
        Chloride                                    &   Numerical                   \\
        Creatinine Blood                            &   Numerical                   \\
        Glucose (Serum)                             &   Numerical                   \\
        Hct                                         &   Numerical                   \\
        Hgb                                         &   Numerical                   \\
        INR (International Normalized Ratio)        &   Numerical                   \\
        LDH                                         &   Numerical                   \\
        MCH                                         &   Numerical                   \\
        MCH                                         &   Numerical                   \\
        MCV                                         &   Numerical                   \\
        Magnesium                                   &   Numerical                   \\
        PT (Prothrombin Time)                       &   Numerical                   \\
        PTT                                         &   Numerical                   \\
        Phosphate                                   &   Numerical                   \\
        Platelet Count                              &   Numerical                   \\
        RBC                                         &   Numerical                   \\
        RDW                                         &   Numerical                   \\
        Sodium                                      &   Numerical                   \\
        WBC                                         &   Numerical                   \\
        PO intake                                   &   Numerical                   \\
        Amiodarone                                  &   Numerical                   \\
        D5W                                         &   Numerical                   \\
        Heparin                                     &   Numerical                   \\
        Famotidine                                  &   Numerical                   \\
        Dextrose Other                              &   Numerical                   \\
        KCl                                         &   Numerical                   \\
        SG Urine                                    &   Numerical                   \\
        pH Urine                                    &   Numerical                   \\
        Fresh Frozen Plasma                         &   Numerical                   \\
        Albumin 5\%                                 &   Numerical                   \\
        Bilirubin (Direct)                          &   Numerical                   \\
        Bilirubin (Indirect)                        &   Numerical                   \\
        Jackson-Pratt                               &   Numerical                   \\
        Albumin                                     &   Numerical                   \\
        Neosynephrine                               &   Numerical                   \\
        Propofol                                    &   Numerical                   \\
        Unknown                                     &   Numerical                   \\
        EBL                                         &   Numerical                   \\
        OR/PACU Crystalloid                         &   Numerical                   \\
        Intubated                                   &   Numerical                   \\
        Stool                                       &   Numerical                   \\
        Gastric                                     &   Numerical                   \\
        Gastric Meds                                &   Numerical                   \\
        Pre-admission Intake                        &   Numerical                  \\
        Pre-admission Output                        &   Numerical                   \\
        Basophils                                   &   Numerical                   \\
        Eoisinophils                                &   Numerical                   \\
        Lymphocytes                                 &   Numerical                   \\
        Monocytes                                   &   Numerical                   \\
        Neutrophils                                 &   Numerical                   \\
        Nitroglycerine                              &   Numerical                   \\
        Chest Tube                                  &   Numerical                   \\
        Packed RBC                                  &   Numerical                   \\
        Colloid                                     &   Numerical                   \\
        Insulin Regular                             &   Numerical                   \\
        Pantoprazole                                &   Numerical                   \\
        Hydromorphone                               &   Numerical                   \\
        Emesis                                      &   Numerical                   \\
        Insulin Humalog                             &   Numerical                   \\
        Insulin largine                             &   Numerical                   \\
        Furosemide                                  &   Numerical                   \\
        Lactated Ringers                            &   Numerical                   \\
        Morphine Sulfate                            &   Numerical                   \\
        Glucose (Whole Blood)                       &   Numerical                   \\
        Calcium Gluconate                           &   Numerical                   \\
        Metoprolol                                  &   Numerical                   \\
        Norepinephrine                              &   Numerical                   \\
        Vasopressin                                 &   Numerical                   \\
        Dopamine                                    &   Numerical                   \\
        Fentanyl                                    &   Numerical                   \\
        Midazolam                                   &   Numerical                   \\
        Creatinine Urine                            &   Numerical                   \\
        Piggyback                                   &   Numerical                   \\
        Magnesium Sulfate (Bolus)                   &   Numerical                   \\
        Magnesium Sulphate                          &   Numerical                   \\
        KCl (Bolus)                                 &   Numerical                   \\
        Nitroprusside                               &   Numerical                   \\
        Lorazepam                                   &   Numerical                   \\
        Piperacillin                                &   Numerical                   \\
        Fiber                                       &   Numerical                   \\
        Residual                                    &   Numerical                   \\
        Free Water                                  &   Numerical                   \\
        GT Flush                                    &   Numerical                   \\
        Vacomycin                                   &   Numerical                   \\
        Hydralazine                                 &   Numerical                   \\
        Half Normal Saline                          &   Numerical                   \\
        Cefazolin                                   &   Numerical                   \\
        Sterile Water                               &   Numerical                   \\
        Ultrafiltrate                               &   Numerical                   \\
        TPN                                         &   Numerical                   \\
        Albumin 25\%                                &   Numerical                   \\
        Epinephrine                                 &   Numerical                   \\
        Milrinone                                   &   Numerical                   \\
        Insulin NPH                                 &   Numerical                   \\
        Lymphocytes (Absolute)                      &   Numerical                   \\
        Temperature                                 &   Numerical                   \\
        Height                                      &   Numerical                   \\
        GCS\_eye (Glasgow Coma Scale Eye Opening)    &   Categorical                 \\              
        GCS\_motor (Glasgow Coma Scale Motor Response)           &   Categorical                 \\
        GCS\_verbal (Glasgow Coma Acale Verbal Response)          &   Categorical                 \\  \bottomrule
    
    \caption{\textbf{Feature in MIMIC-III Dataset.}}
    \label{table:mi3_feature}
    \end{longtable}

\begin{table}[htp]
    \centering
    \resizebox{0.7\linewidth}{!}{
    \begin{tabular}{ccc}
    \toprule
    \centering
    \textbf{Feature}                                &   \textbf{Feature Type}       \\  \midrule
        Weight                                      &   Numerical                   \\
        ALP (Alkaline Phosphatase)                  &   Numerical                   \\
        ALT (Alanine Aminotransferase)              &   Numerical                   \\
        AST (Aspartate Aminotransferase)            &   Numerical                   \\
        ALB (Albumin)                               &   Numerical                   \\
        BUN (Blood Urea Nitrogen)                   &   Numerical                   \\
        Bilirubin                                   &   Numerical                   \\
        Cholesterol                                 &   Numerical                   \\
        Creatinine                                  &   Numerical                   \\
        DiasABP (Diastolic Arterial Blood Pressure) &   Numerical                   \\
        FiO2 (Inspired Fraction of Oxygen)          &   Numerical                   \\
        GCS (Glasgow Coma Scale)                    &   Categorical                 \\
        Glucose                                     &   Numerical                   \\
        HCO3 (Bicarbonate)                          &   Numerical                   \\
        HCT (Hematocrit)                            &   Numerical                   \\
        HR (Heart Rate)                             &   Numerical                   \\
        K (Potassium)                               &   Numerical                   \\
        Lactate                                     &   Numerical                   \\
        MAP (Mean Arterial Pressure)                &   Numerical                   \\
        MechVent (Mechanical Ventilation)           &   Categorical                 \\
        Mg (Magnesium)                              &   Numerical                   \\
        PaCO2 (Partial Pressure of Carbon Dioxide)  &   Numerical                   \\
        PaO2 (Partial Pressure of Oxygen)           &   Numerical                   \\
        PLT (Platelets)                             &   Numerical                   \\
        RespRate (Respiratory Rate)                 &   Numerical                   \\
        SaO2 (Arterial Oxygen Saturation)           &   Numerical                   \\
        SysABP (Systolic Arterial Blood Pressure)   &   Numerical                   \\
        Temp (Temperature)                          &   Numerical                   \\
        TroponinI                                   &   Numerical                   \\
        TroponinT                                   &   Numerical                   \\
        Urine                                       &   Numerical                   \\
        WBC (White Blood Cell)                      &   Numerical                   \\
        pH (Body Fluid)                             &   Numerical                   \\
        Age                                         &   Numerical                   \\
        Height                                      &   Numerical                   \\
        Gender                                      &   Categorical                 \\
        ICU Type                                    &   Categorical                 \\  \bottomrule
    \end{tabular}
    }
    \vspace*{0.2cm}
    \caption{\textbf{Feature in PhysioNet2012 Dataset.}}
    \label{table:p12_feature}
\end{table}
\clearpage

\begin{table}[htp]
    \centering
    \resizebox{0.7\linewidth}{!}{
    \begin{tabular}{ccc}
    \toprule
    \centering
    \textbf{Feature}                                &   \textbf{Feature Type}       \\  \midrule
        AFP (Alpha-Fetoprotein)                     &   Numerical                   \\
        ALB (Albumin)                               &   Numerical                   \\
        ALP (Alkaline Phosphatase)                  &   Numerical                   \\
        ALT (Alanine Aminotransferase)              &   Numerical                   \\
        AST (Aspartate Aminotransferase)            &   Numerical                   \\
        Anti-HBc (Hepatitis B Core Antibody)        &   Categorical                 \\
        Anti-HBe (Anti-Hepatitis B e-Antigen)       &   Categorical                 \\
        Anti-HBs (Hepatitis B Surface Antibody)     &   Categorical                 \\
        Anti-HCV (Anti-Hepatitis C Virus Antibody)  &   Categorical                 \\
        BUN (Blood Urea Nitrogen)                   &   Numerical                   \\
        CRE (Creatinine)                            &   Numerical                   \\
        D-BIL (Direct Bilirubin)                    &   Numerical                   \\
        GGT (gamma-Glutamyltransferase)             &   Numerical                   \\
        Glucose AC                                  &   Numerical                   \\
        HB (Hemoglobin)                             &   Numerical                   \\
        HBVDNA (Hepatitis B Virus DNA)              &   Numerical                   \\
        HBeAg (Hepatitis B e-Antigen)               &   Categorical                 \\
        HBsAg (Hepatitis B Surface Antigen)         &   Categorical                 \\
        HCVRNA (Hepatitis C Virus RNA)              &   Numerical                   \\
        HbA1c (Glycated Haemoglobin)                &   Numerical                   \\
        Lym (Lymphocyte)                            &   Numerical                   \\
        Na (Sodium)                                 &   Numerical                   \\
        PLT (Platelet)                              &   Numerical                   \\
        PT (Prothrombin Time)                       &   Numerical                   \\
        PT INR (PT International Normalized Ratio)  &   Numerical                   \\
        Seg (Neutrophils)                           &   Numerical                   \\
        T-BIL (Total Bilirubin)                     &   Numerical                   \\
        TP (Total Protein)                          &   Numerical                   \\
        WBC (White Blood Cell)                      &   Numerical                   \\
        Height                                      &   Numerical                   \\
        Weight                                      &   Numerical                   \\
        fatty\_liver                                &   Categorical                 \\
        paranchymal\_liver\_disease                 &   Categorical                 \\
        Age                                         &   Numerical                   \\
        hosp\_days                                  &   Numerical                   \\
        Sex                                         &   Categorical                 \\
        sono                                        &   Categorical                 \\  \bottomrule
    \end{tabular}
    }
    \vspace*{0.2cm}
    \caption{\textbf{Feature in Anonymous Hepatocellular Carcinoma Dataset.}}
    \label{table:ntuh_feature}
\end{table}

\clearpage

\section{Model Architecture and Hyperparameters}
\label{appendix:grid_search}

For Transformer Encoder and SUMMIT, they are composed of two modules. One is a feature extractor, and the other is a single classifier to predict the probability of classes. The classifier module contains a dense layer and a linear layer. The dense layer contains a linear layer and an activation function $\mathrm{GELU}$ \citep{hendrycks_gaussian_2023}. The sequence data for the Transformer Encoder is concatenated with the missing mask for each timestamp. We aggregate the outputs from the Transformer-based feature extractor module by their means.

For Random Forest and XGBoost, we employ their empirically optimal default hyperparameters. The data for these two models is subjected to a \emph{summarization} strategy, wherein the missing mask is concatenated to the original data by timestamps. Before being fed to the model, data is flattened into a one-dimensional matrix. 

For the SOTA deep models, we adopt the best settings reported in the original papers if available.

\begin{table*}[htp]
    \centering
    \resizebox{0.9\columnwidth}{!}{%
    \begin{tabular}{ccc}
    \toprule
    \centering
    \textbf{Hyperparameter}                                 &      \textbf{Transformer Encoder}        &   \textbf{SUMMIT}      \\  \midrule
    d\_model                                                                   &   112, 128, 144, 160                  &   112, 128, 144, 160      \\  \midrule
    num\_head                                                                  &   1                                   &   1, 2                       \\  \midrule
    ff\_dim                                                                    &   64, 80, 96, ..., 240, 256           &   64, 80, 96, ..., 240, 256\\  \midrule
    hidden\_size                                            &   --                                   &   --                       \\  \midrule
    num\_layer                                              &   1, 2, 4, 8, 16                      &   1, 2, 4, 8              \\  \midrule
    classifier\_down\_factor                                                   &   2                                   &   2, 4, 6, 8                       \\  \midrule
    learning rate                                               &   3e-3, 3e-4, 3e-5                    &   3e-3, 3e-4, 3e-5        \\  \midrule
    Optimizer                                                              &   Adam                                &   Adam                    \\  \bottomrule
    \end{tabular}
    }
    \caption{\textbf{Grid Searching Range of All Deep Learning Models}}
    \label{tab:grid_search_range}

\hspace*{\fill}

    \resizebox{\columnwidth}{!}{%
    \begin{tabular}{ccllllc}\toprule
    \centering
    \textbf{Hyperparameter}                                        &   \textbf{TCN}& \textbf{SAnD }& \textbf{STraTS }&\textbf{mTAN } &\textbf{SUMMIT}      
&   \textbf{Transformer}\\  \midrule
    d\_model                                                                   &   --                    & --& --&256 &144                     
&   128                                 \\  \midrule
    ff\_dim                                                                    &   --                    & --& --&20 &80                      
&   144                                 \\  \midrule
    hidden\_size                                                             &   128& 64& 64&-- &--                       
&   --                                   \\  \midrule
    num\_layer                                                                &   4& 4& 2&1 &1                       
&   8                                   \\  \midrule
    learning rate                                                         &   5e-4& 5e-4& 5e-4&5e-5 &3e-5                    
&   3e-5                                \\  \midrule
    early stopping epoch                                                    &   23& 25& 50&200 &380                     &   125                                 \\  \bottomrule
    \end{tabular}%
    }
    \caption{\textbf{The Setting of Hyperparameter in MI3 Dataset}}
    \label{tab:setting_mi3}

\hspace*{\fill}

    \resizebox{\columnwidth}{!}{%
    \begin{tabular}{ccllllc}\toprule
    \centering
    \textbf{Hyperparameter}                                        &   \textbf{TCN} &\textbf{SAnD } & \textbf{STraTS }&\textbf{mTAN } &\textbf{SUMMIT}      
&   \textbf{Transformer}\\  \midrule
    d\_model                                                                   &   --                    & --& --&256 &144                     
&   144                                 \\  \midrule
    ff\_dim                                                                    &   --                    & --& --&20 &144                     
&   144                                 \\  \midrule
    hidden\_size                                                            &   64& 64& 64&-- &--                       
&   --                                   \\  \midrule
    num\_layer                                                                 &   4& 4& 2&1 &1                       
&   8                                   \\  \midrule
    learning rate                                                           &   5e-4& 5e-4& 5e-4&1e-4 &3e-5                    
&   3e-5                                \\  \midrule
    early stopping epoch                                                    &   23& 30& 50&250 &350                     &   75                                  \\  \bottomrule
    \end{tabular}%
    }
    \caption{\textbf{The Setting of Hyperparameter in P12 Dataset}} 
    \label{tab:setting_p12}

\hspace*{\fill}

    \resizebox{\columnwidth}{!}{%
    \begin{tabular}{ccllllc}\toprule
    \centering
    \textbf{Hyperparameter}                                         &   \textbf{TCN} & \textbf{SAnD }& \textbf{STraTS }&\textbf{mTAN } &\textbf{SUMMIT}      
&   \textbf{Transformer}\\  \midrule
    d\_model                                                                   &   --                    & --& --&256 &144
&   144                                 \\  \midrule
    ff\_dim                                                                    &   --                    & --& --&20 &144                     
&   144                                 \\  \midrule
    hidden\_size                                                             &   64& 64& 64&-- &--                       
&   --                                   \\  \midrule
    num\_layer                                                                &   4& 4& 2&1 &8                       
&   16                                  \\  \midrule
    learning rate                                                           &   5e-4& 5e-4& 5e-4&1e-4 &3e-5&   3e-5                                \\  \midrule
    early stopping epoch                                                     &   75                   & 29& 44&54 &100                     &   85                                  \\  \bottomrule
    \end{tabular}%
    }
    \caption{\textbf{The Setting of Hyperparameter in HCC Dataset}}
    \label{tab:setting_hc2}

\end{table*}

\cref{tab:grid_search_range} shows the hyperparameter grid searching range of all deep learning models. "d\_model" means the dimension of embeddings. "num\_head" is the number of attention heads. "ff\_dim" is the feed-forward dimension of attention module in the transformer-based models. "hidden\_size" is the dimension of hidden vectors in the GRU-based models. "num\_layer" is the number of unit stacks. We use Adam optimizer \citep{kingma_adam_2017} to optimize all models. After the grid searching, we will do a little perturbation on the learning rate to see if the model performs better. Restricted by the GPU memory, the "num\_layer" of SUMMIT on the PhysioNet2012 dataset is set to 1. We also list all settings of all models in \cref{tab:setting_mi3}, \ref{tab:setting_p12}, and \ref{tab:setting_hc2}. 

\section{Platform Information}
\label{appendix:plf_info}

The following is the information on the main environment we used to conduct all the experiments in the paper.

\begin{itemize}
  \item{Hardware-1:} 
  \begin{itemize}
    \item{CPU: Intel(R) Core(TM) i7-10700 CPU @ 2.90GHz}
    \item{Memory: 64GB}
    \item{GPU: RTX 3060 with 12GB VRAM}
  \end{itemize}
  
  \item{Hardware-2:} 
  \begin{itemize}
    \item{CPU: Intel(R) Xeon(R) Gold 6154 CPU @ 3.00GHz}
    \item{Memory: 56GB}
    \item{GPU: Tesla V100 PCle with 32GB VRAM}
  \end{itemize}

  \item{Platform:} 
  \begin{itemize}
    \item{CUDA version: 11.4 / 12.3}
    \item{gcc version: 7.5.0 / 11.4.0}
    \item{pytorch version: 1.13.1}
    \item{sklearn version: 1.1.2}
    \item{xgboost version: 1.7.5}
  \end{itemize}
\end{itemize}

\clearpage

\section{Masking}
In this section, we retrain and retest our model, SUMMIT, on these three datasets without masking missing values. The missing values here are imputed with the global mean and the global mode of the training set. The main reason for this ablation study is to check if imputation is useless for our model. Due to the class imbalance, we take AUPRC as the metric. \cref{tab:masking_comparison} shows that our model performs better with masking the missing on each dataset. This may imply the imputation confuses the model when training and testing, and the imputation in these three datasets may not be a good solution for missing values.

\begin{table}[htp]
    \centering
    \resizebox{0.6\linewidth}{!}{
    \begin{tabular}{cccc} \toprule
    \textbf{Dataset}  &  \makecell{\textbf{HCC}} & \textbf{P12}& \textbf{MI3} \\ \midrule
    w/o Masking Missing & 0.4180 & 0.5283 & 0.5679 \\ \midrule
    w/ Masking Missing & 0.4553 & 0.5504 & 0.6328 \\ \bottomrule
    \end{tabular}%
    }
    \vspace*{0.5cm}
    \caption{\textbf{AUPRC of SUMMIT w/ or w/o Masking Missing}: The settings of these two SUMMITs are identical, including the hyperparameters, to ensure an identical model architecture. The only difference is whether to mask imputation when training and testing.}
    \label{tab:masking_comparison}
\end{table}

\end{document}